\documentclass[journal]{IEEEtran}

\ifCLASSINFOpdf
\else
   \usepackage[dvips]{graphicx}
\fi
\usepackage{url}

\usepackage{float} 
\usepackage[colorlinks,
linkcolor=blue,
anchorcolor=blue,
citecolor=blue]{hyperref}
\usepackage{graphicx}
\usepackage{enumitem}
\usepackage{orcidlink}
\usepackage{multirow}
\usepackage{makecell}
\begin{document}

\title{A Visual Question Answering Method for SAR Ship: Breaking the Requirement for Multimodal Dataset Construction and Model Fine-Tuning}

\author{Fei Wang, Chengcheng Chen,~\IEEEmembership{Graduate Student Member, IEEE}, \\
Hongyu Chen, Yugang Chang, and Weiming Zeng,~\IEEEmembership{Senior Member, IEEE}
\thanks{This paragraph of the first footnote will contain the date on which you submitted your paper for review. It will also contain support information, including sponsor and financial support acknowledgment. (Corresponding author: Weiming Zeng)}
\thanks{The authors are with the Digital Imaging and Intelligent Computing Laboratory, Shanghai Maritime University, Shanghai 201306, China (e-mail:shine\_wxf@163.com; shmtu\_ccc@163.com; hongychen676@gmail.com; 202340310001@shmtu.edu.cn; zengwm86@163.com)}}

\markboth{Journal of \LaTeX\ Class Files, Vol. 14, No. 8, August 2015}
{Shell \MakeLowercase{\textit{et al.}}: Bare Demo of IEEEtran.cls for IEEE Journals}
\maketitle

\begin{abstract}
Current visual question answering (VQA) tasks often require constructing multimodal datasets and fine-tuning visual language models, which demands significant time and resources. This has greatly hindered the application of VQA to downstream tasks, such as ship information analysis based on Synthetic Aperture Radar (SAR) imagery. To address this challenge, this letter proposes a novel VQA approach that integrates object detection networks with visual language models, specifically designed for analyzing ships in SAR images. This integration aims to enhance the capabilities of VQA systems, focusing on aspects such as ship location, density, and size analysis, as well as risk behavior detection. Initially, we conducted baseline experiments using YOLO networks on two representative SAR ship detection datasets, SSDD and HRSID, to assess each model's performance in terms of detection accuracy. Based on these results, we selected the optimal model, YOLOv8n, as the most suitable detection network for this task. Subsequently, leveraging the vision-language model Qwen2-VL, we designed and implemented a VQA task specifically for SAR scenes. This task employs the ship location and size information output by the detection network to generate multi-turn dialogues and scene descriptions for SAR imagery. Experimental results indicate that this method not only enables fundamental SAR scene question-answering without the need for additional datasets or fine-tuning but also dynamically adapts to complex, multi-turn dialogue requirements, demonstrating robust semantic understanding and adaptability. 
\end{abstract}

\IEEEpeerreviewmaketitle

\section{Introduction}
\IEEEPARstart{I}{n} recent years, deep learning has driven major advancements in object detection, natural language processing, and large language models\cite{10555327}. Object detection models analyze images to provide information about the location and size of objects within them. Natural language processing models respond to user input and requirements to generate answers. Visual Question Answering (VQA) systems attempt to interpret images by combining image data with prompts and leveraging their general knowledge to respond to users\cite{10734181}. Recently, VQA has gained attention in the fields of Geoscience and remote sensing, with studies such as \cite{10500421},\cite{9444570} advancing the application of VQA for remote sensing information. As a form of remote sensing imagery, Synthetic Aperture Radar (SAR) offers all-weather monitoring capabilities unaffected by weather conditions\cite{10518076}, holding significant promise for maritime traffic applications.

You Only Look Once (YOLO) \cite{Li2022YOLOv6AS}, \cite{10204762}, \cite{Wang2024YOLOv10RE}, \cite{Jocher_Ultralytics_YOLO_2023} excels in object detection with fast and accurate results, making it effective for various tasks, though its applications are relatively limited. Vision Language Models (VLMs) like those used in VQA typically require extensive data and training to achieve good performance, particularly when applied to specialized tasks such as SAR imagery in remote sensing. Parameter-efficient fine-tuning (PEFT) \cite{peft} with methods like LoRA \cite{Hu2021LoRALA} is often necessary but demands considerable resources and high-performance hardware. To ensure efficiency, lower-parameter models are preferred but generally lack broad scientific knowledge. Alternatively, multimodal fusion networks can integrate image and text to answer questions; however, approaches like those by Lobry et al. \cite{9088993} and Li et al. \cite{10500421} that use linear layers for classification are constrained to predefined question types and struggle with complex, open-ended queries.

Researchers have developed various vision-language models, like Qwen2-VL \cite{Wang2024Qwen2VLEV} and BLIP-2 \cite{Li2023BLIP2BL}, which possess extensive general knowledge and are well-suited for VQA tasks without the need for additional fine-tuning or dataset creation. However, SAR imagery poses unique challenges due to its high noise levels that can interfere with the model's understanding, and the often unclear ship features, particularly in nearshore scenes or when dealing with small targets. Additionally, although these vision-language models recognize image content well, they struggle to extract specific details such as target location and size, which are essential for accurately addressing user queries.

Based on the background outlined above, this letter presents a novel visual question answering approach that focuses on rapidly transferring visual language question answering tasks to the maritime traffic domain using SAR imagery, without the need for additional dataset construction or model fine-tuning. The specific contributions are as follows:
\begin{enumerate}[label=\arabic*)]
	\item We propose an efficient method for ship-related querying in SAR images by connecting an object detection network with a vision-language model, thereby eliminating the need for model fine-tuning or multimodal dataset construction.
	
	\item Baseline experiments using YOLO series models were conducted on the SSDD and HRSID datasets to systematically assess each model’s accuracy and efficiency in SAR ship detection. The optimal model was selected to construct prompts based on predicted bounding box information. Q\&A experiments on representative samples demonstrate that the proposed method enables more accurate and reliable responses, offering new insights for future multimodal SAR image analysis and maritime traffic monitoring.
\end{enumerate}
\section{METHODOLOGY}
\begin{figure}
	\centerline{\includegraphics[width=0.65\columnwidth]{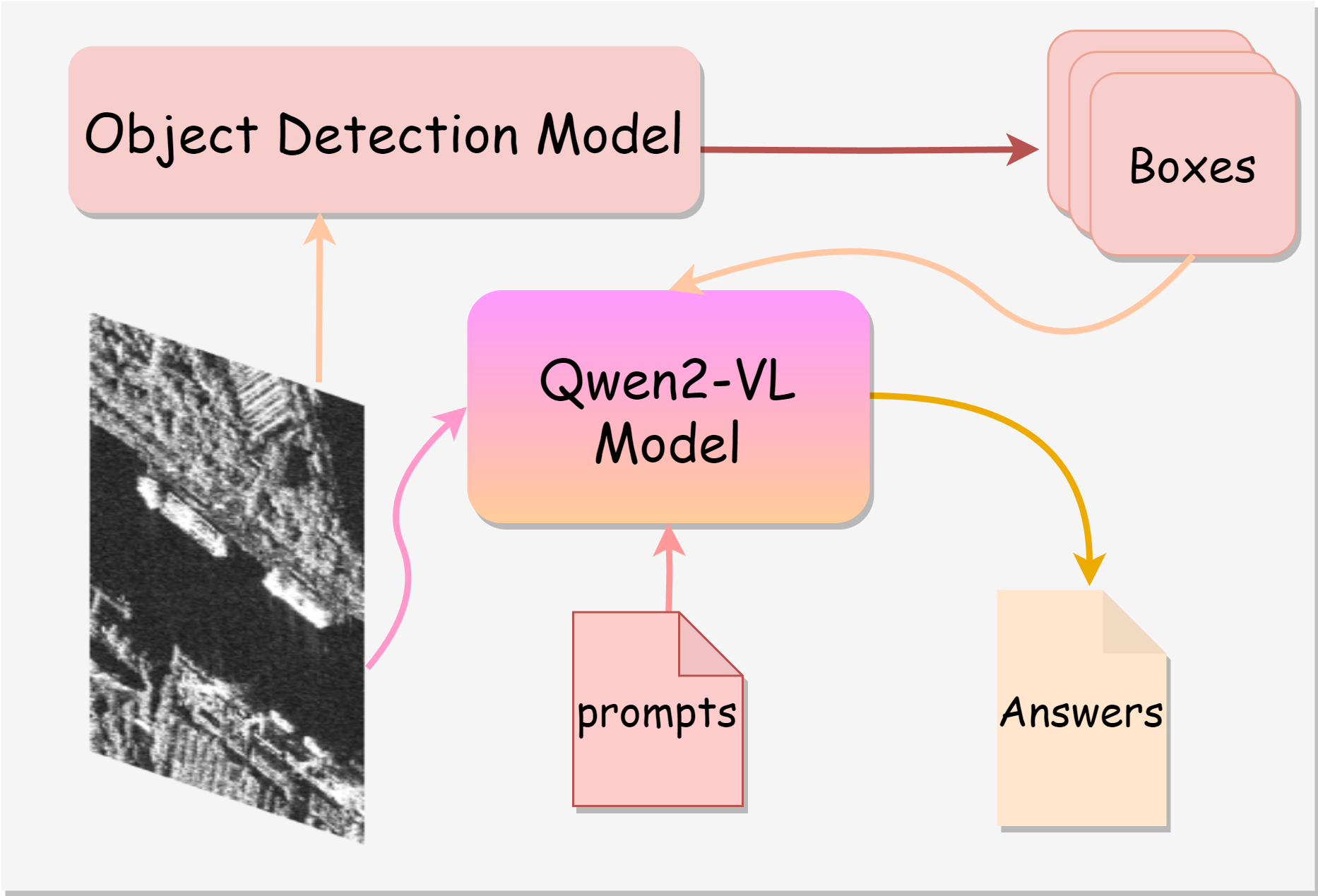}}
	\caption{Overview of the proposed method.}
	\label{fig1}
\end{figure}

The proposed method encompasses two primary components: an Object Detection Model (ODM) and a Vision-Language Model (VLM). At the initiation of the Q\&A process, the image is first processed by the ODM to detect and extract bounding boxes (Boxes) for the ships. Subsequently, these Boxes are integrated with the user's prompts to formulate a composite text. This combined text, along with the image, is then fed into the VLM. The VLM generates a response to the user’s query, and both the user’s questions and the VLM’s responses are logged, allowing for multi-turn conversations. The overall architecture of the proposed method is illustrated in Fig. \ref{fig1}.

\subsection{ODM}
Analyzing maritime and nearshore vessel traffic information based on SAR imagery requires the extraction of vessel location and size, which is crucial. In visual question answering (VQA) tasks, users often need to quickly access the information they seek, necessitating a high-performance detection module. The YOLO series algorithms are renowned for their exceptional detection accuracy and efficiency. Therefore, this letter selects the YOLO series as the feature extractor, with YOLOv8n chosen as the object detection model based on experimental results.

\subsection{VLM}

The VLM serves as the cornerstone of the VQA task, responsible for interpreting image information and integrating user input to generate responses. In this study, we utilize the Qwen2-VL model, an open-source model developed by Alibaba Group. Qwen2-VL offers versions with varying parameter sizes; we select Qwen2-VL-72B to ensure the model possesses extensive general knowledge in social sciences, enhancing robustness for the Q\&A tasks. 

\section{EXPERIMENT}
To validate the effectiveness of the proposed method, we conducted experiments in two phases. First, we trained and evaluated the YOLO series detection algorithms on the SSDD and HRSID datasets. Subsequently, we tested the combination of the trained detection model and the vision-language model on selected representative samples. We posed several domain-specific questions related to maritime traffic to evaluate the accuracy of the model's responses.

\subsection{Datasets}
Our experiments are primarily based on the SSDD\cite{rs13183690} and HRSID\cite{9127939} datasets. The SSDD dataset comprises a total of 1160 images, which are divided according to the dataset’s paper: 928 images for training and 232 for testing. The HRSID dataset, with fixed image dimensions, consists of 5604 images, divided according to the official split: 3642 images for training and 1962 for testing.

\subsection{Implementation Details}
Our object detection accuracy experiments were conducted on a Linux system using an Intel Xeon Silver 4210R CPU and four Nvidia GeForce RTX 2080Ti GPUs. Detailed experimental settings are provided in Table \ref{table2}. After training, we evaluated the performance metrics on the test set. In this study, we used mean Average Precision at IoU = 0.5 (mAP.50) and mean Average Precision at IoU thresholds ranging from 0.5 to 0.95 in increments of 0.05 (mAP.50:.95) as the primary metrics. The best-performing model was then selected to extract ship information, and custom Python scripts were developed to combine user input with extracted ship details. This combined information, along with the original image, was input to the VQA model for multi-turn dialogue interactions.

\begin{table}[h!]
	
	\begin{center}
		\caption{Specific Experimental Details.}
		\label{table2}
		\setlength{\tabcolsep}{1mm}{
			\begin{tabular}{c c c c c c c}
				\hline
				Dataset & Epochs & Optimizer & Batch & LR0 &  Momentum & Weight\_decay\\
				\hline
				SSDD & 300 & AdamW & 16 & 0.002 & - &5e-4\\
				HRSID & 300 & SGD & 16 & 0.01  & 0.9 &5e-4\\
				\hline
			\end{tabular}
		}
	\end{center}

\end{table}

\subsection{ODM selection}
The experimental results are shown in Table \ref{table3}. Based on the model's overall performance across both datasets, YOLOv8n was selected as the optimal ODM.
\begin{table}[h!]

	\begin{center}
		\caption{ODM Selection Experiments on SSDD and HRSID.}
		\label{table3}
		\setlength{\tabcolsep}{4.5pt}
		\begin{tabular}{c c c c c}
			\Xhline{2\arrayrulewidth}
			\multirow{2}*{Method} & \multicolumn{2}{c}{SSDD} & \multicolumn{2}{c}{HRSID} \\
			
			~  &  mAP\textsubscript{.50}(\%) & mAP\textsubscript{.50:.95}(\%) & mAP\textsubscript{.50}(\%) & mAP\textsubscript{.50:.95}(\%) \\
			\hline
			\rule{0pt}{1em}YOLOv6n & 96.9 & 71.2 & 88.2 & 62.8\\
			\rule{0pt}{1em}YOLOv7-tiny & 96.4 & 66.5 & 85.4 & 57.2\\
			\rule{0pt}{1em}YOLOv8n & \textbf{98.6} & 73.1 & \textbf{91.3} & \textbf{67.5}\\
			\rule{0pt}{1em}YOLOv10n &  96.8 & 72.6 & 90.3 & 66.9\\
			\rule{0pt}{1em}YOLO11n & 98.0 & \textbf{73.2} & 90.4  & 66.9\\
			\Xhline{2\arrayrulewidth}
			
		\end{tabular}
	\end{center}
\end{table}

\subsection{VQA EXPERIMENT}
\subsubsection{Experimental Design}
\begin{figure}
	\centerline{\includegraphics[scale=0.28]{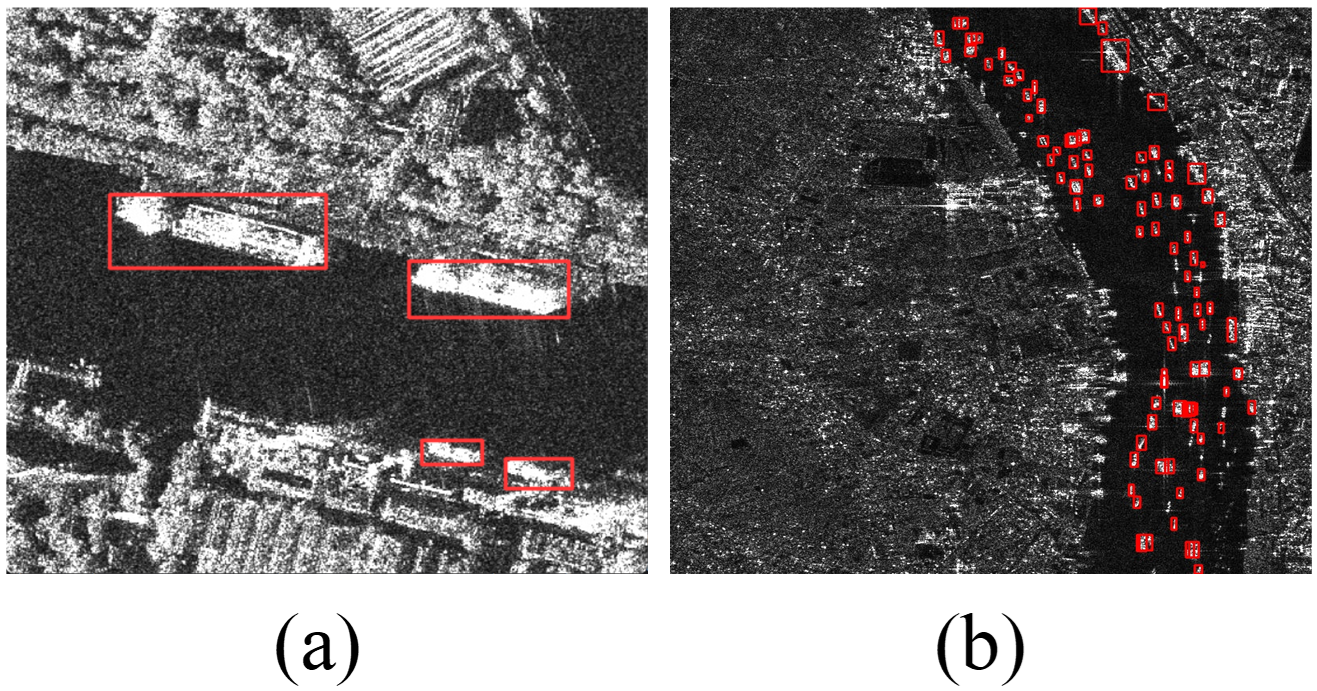}}
	\caption{Samples with Prediction Boxes}
	\label{fig3}
\end{figure}

We selected several representative samples from the SSDD and HRSID test sets and designed a series of specialized maritime traffic questions. Subsequently, we tested the visual language model’s response accuracy in two scenarios: one without providing ship box data and one with the ship box data included.The samples with prediction boxes is shown in Fig. \ref{fig3}, where the column (a) is sample 1 and the column (b) is sample 2. However, in the ensuing dialogue, the images input to the VQA model were raw images, but box information was introduced in the prompts. Since the first conversation involves picture input and a question guide, our dialogue turn numbering starts at 0. 

\begin{figure}
	\centerline{\includegraphics[scale=0.37]{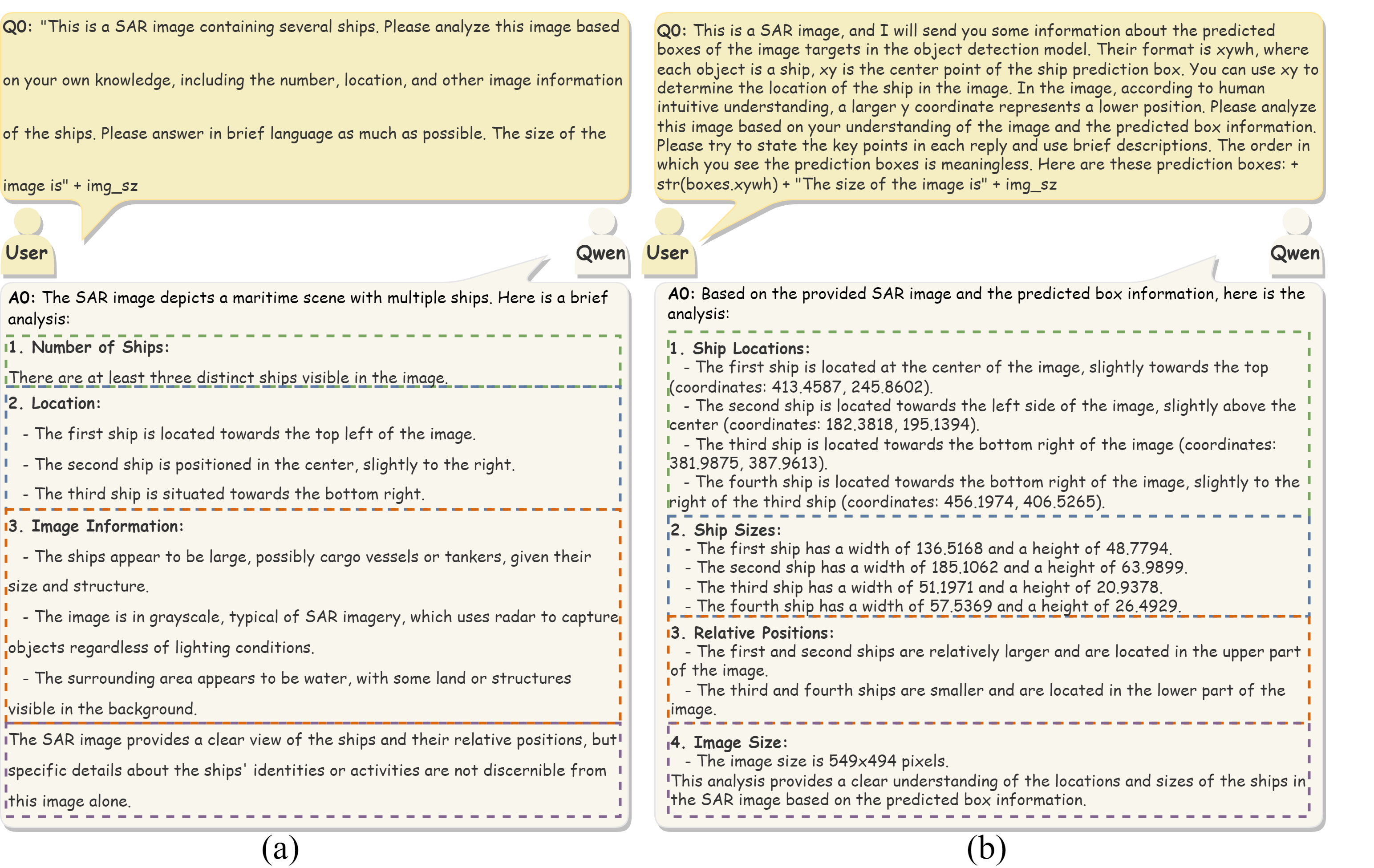}}
	\caption{Sample 1 Dialogue 0.}
	\label{fig4}
\end{figure}
\subsubsection{VQA On Sample 1}
\begin{figure}
	\centerline{\includegraphics[scale=0.37]{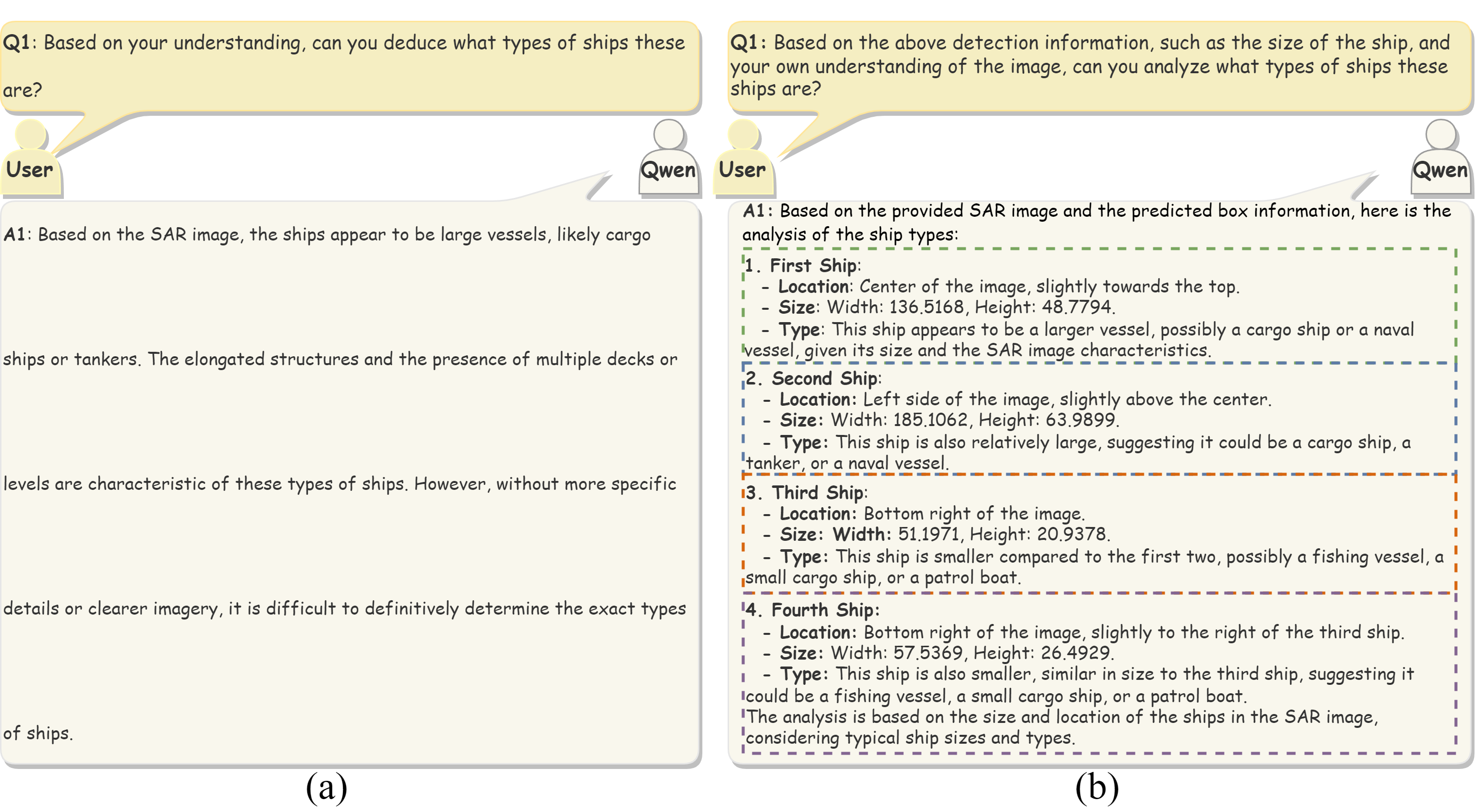}}
	\caption{Sample 1 Dialogue 1.}
	\label{fig5}
\end{figure}

\begin{figure}
	\centerline{\includegraphics[scale=0.37]{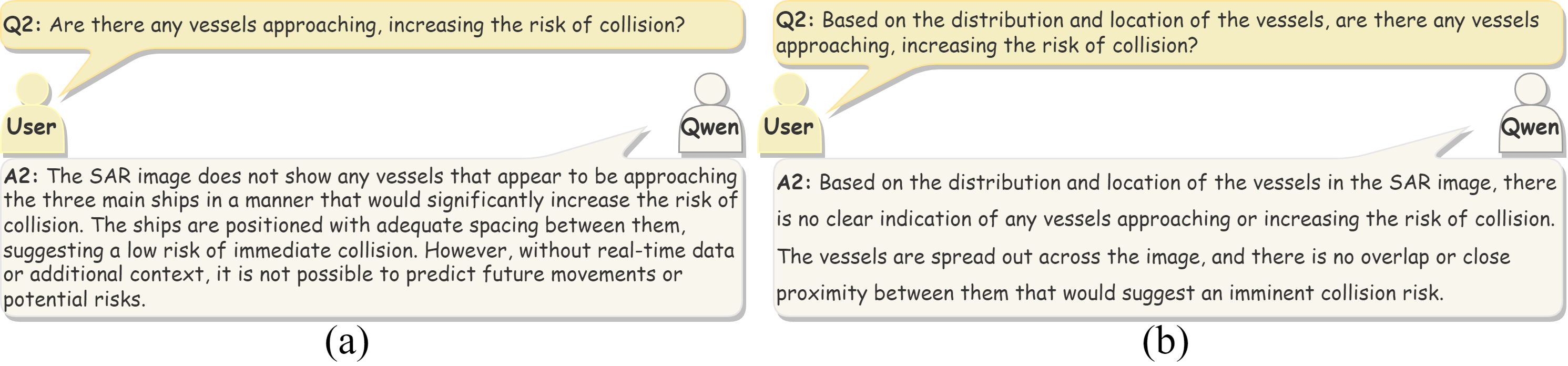}}
	\caption{Sample 1 Dialogue 2.}
	\label{fig6}
\end{figure}

\begin{figure}
	\centerline{\includegraphics[scale=0.37]{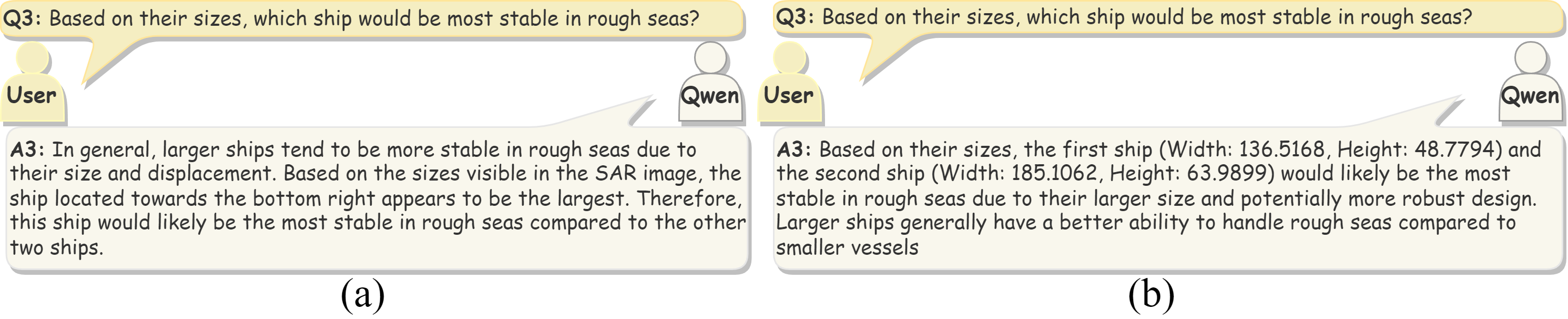}}
	\caption{Sample 1 Dialogue 3.}
	\label{fig7}
\end{figure}

Fig. \ref{fig4} through \ref{fig7} illustrate the dialogues for Sample 1. In these images, subfigure (a) represents user prompts without the predicted ship bounding boxes, while subfigure (b) includes them. As shown in Fig. \ref{fig4}, in Dialogue 0, when the prompts exclude predicted bounding boxes, the VQM cannot extract accurate ship information from the image. Its vague descriptions also make it impossible to locate the corresponding ship within the image.

In Dialogue 1, As shown in Fig. \ref{fig5}, the VQM is asked to infer the type of ships. Without the provision of Boxes, the VQM’s answers do not match the user’s questions due to the lack of precise ship information. In our approach, the VQM inferred the type of each ship by its size.

As shown in Fig. \ref{fig6}, while there are similarities between the two responses in this conversation, the response on the right stands out as more reliable. This is due to the fact that, in the first conversation, the VQM's lack of knowledge about the exact locations of the ships on the image led to a response that was unsupported by specific data.

\begin{figure}
	\centerline{\includegraphics[scale=0.37]{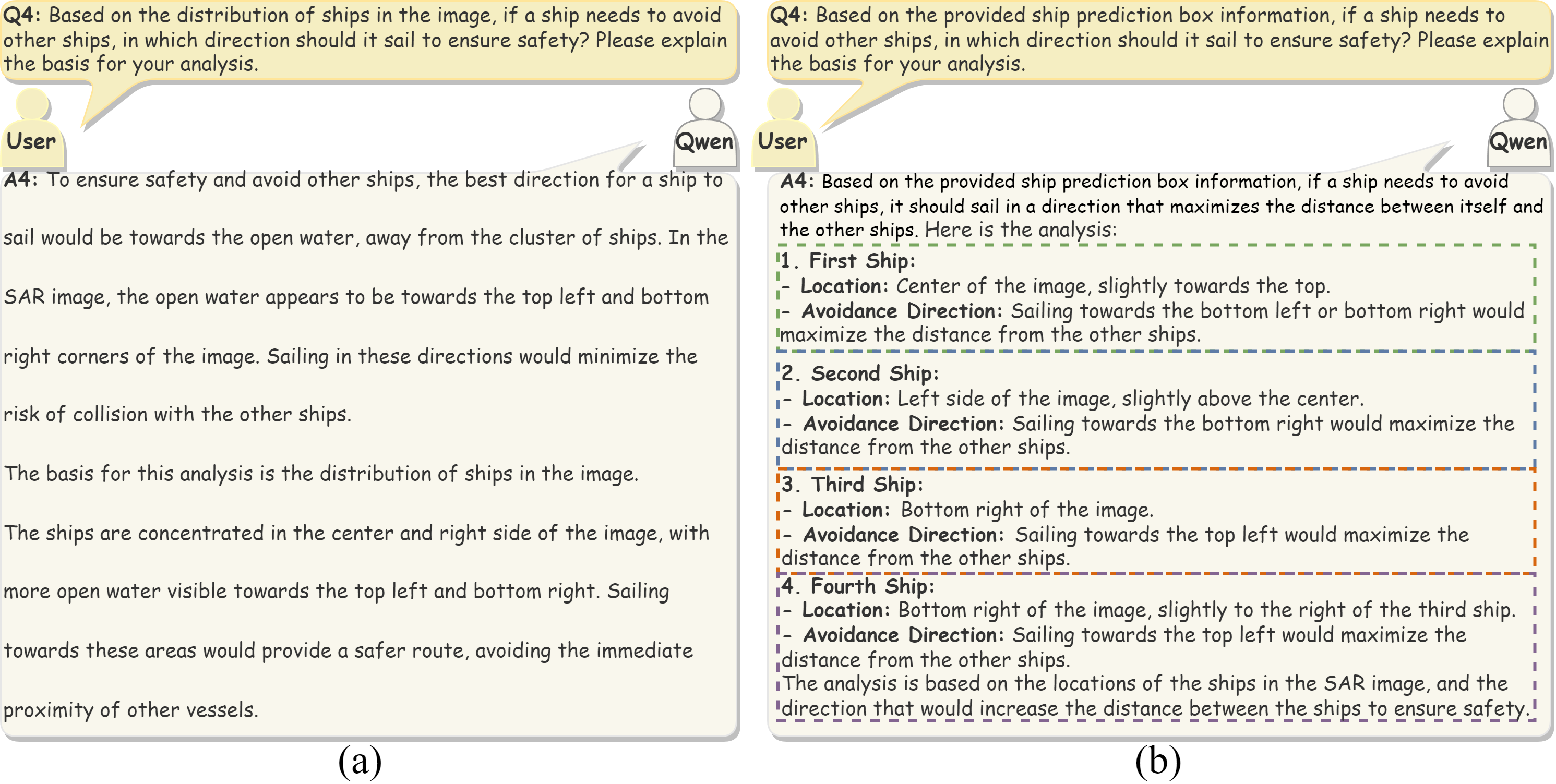}}
	\caption{Sample 1 Dialogue 4.}
	\label{fig8}
\end{figure}

As shown in Fig. \ref{fig7}, in the dialogue on the left, since the VQM does not know the location of the ship, the user will not be able to know which ship it is referring to in its reply; however, after using our method, the VQM reply points to specific data, and therefore the conclusion is more plausible.

In Fig. \ref{fig8}, this conversation is designed as a ship avoidance problem for maritime traffic, asking VQMs to determine the direction of travel to avoid collisions between ships; this problem necessitates not only VQM's understanding of the relative positions of the ships, but also their ability to engage in path planning. In the dialogue on the left of the image, the VQM provides a vague and generic response due to its lack of knowledge about the ships exact positions. After applying our method, the VQM's responses become more precise, and the planning solutions align with the image scenario.
\subsubsection{VQA On Sample 2}
"The second sample represents a complex scenario, which involves the aggregation of small target ships. For this scenario, we designed questions related to maritime transportation and marine safety to evaluate the model's responses. Since the scenario of this test is more complex and the model answers are more detailed, to better demonstrate the effectiveness of incorporating ship prediction information in the responses, we solely present the dialogue information generated by our method.
\begin{figure}
	\centerline{\includegraphics[scale=0.45]{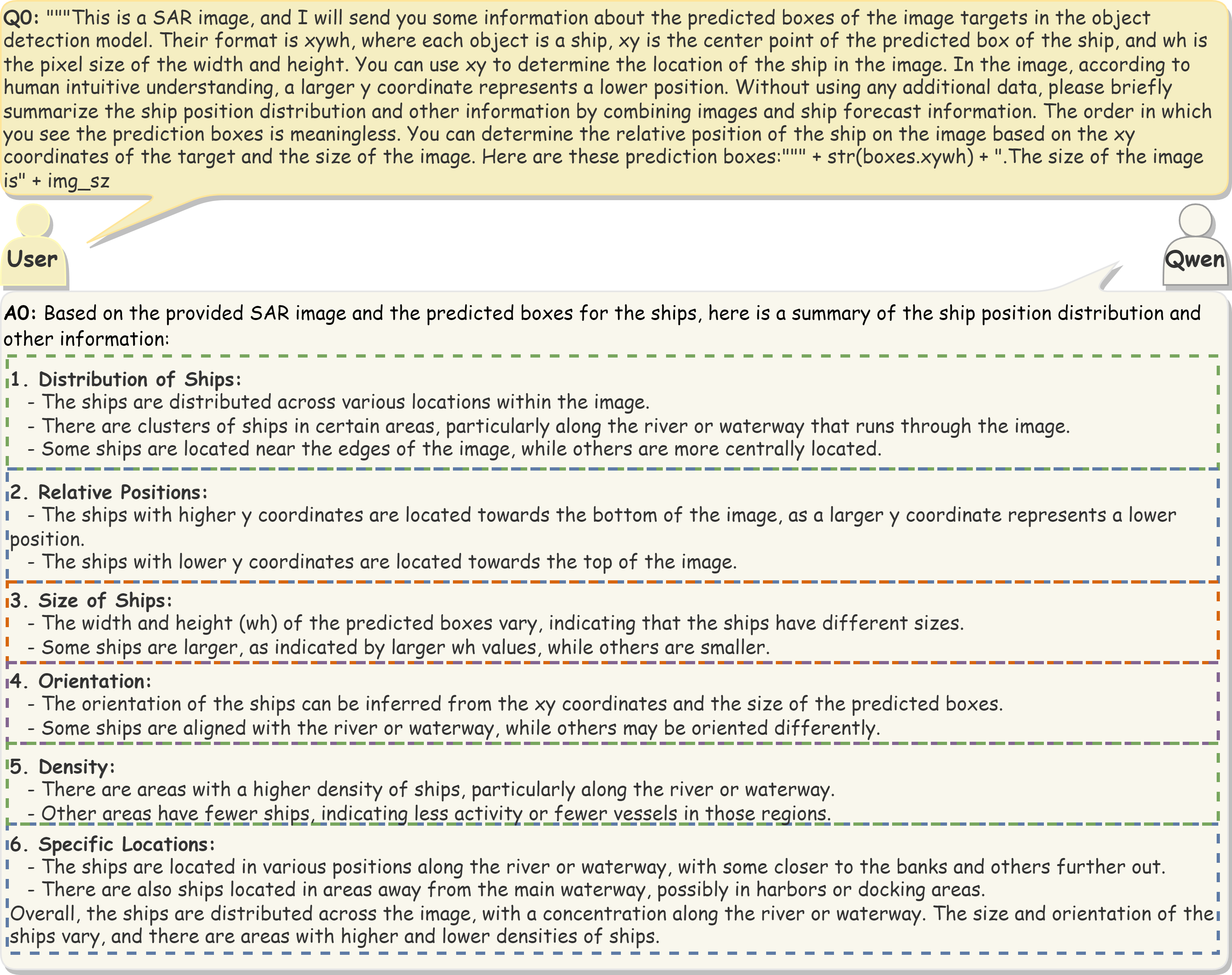}}
	\caption{Sample 2 Dialogue 0.}
	\label{fig9}
\end{figure}

In dialogue 0, as shown in Fig. \ref{fig9}, VQM recognized that the ships were distributed among the river and that the density of ships varied from one area to another, with some of the ships docked along the shoreline.

\begin{figure}
	\centerline{\includegraphics[scale=0.45]{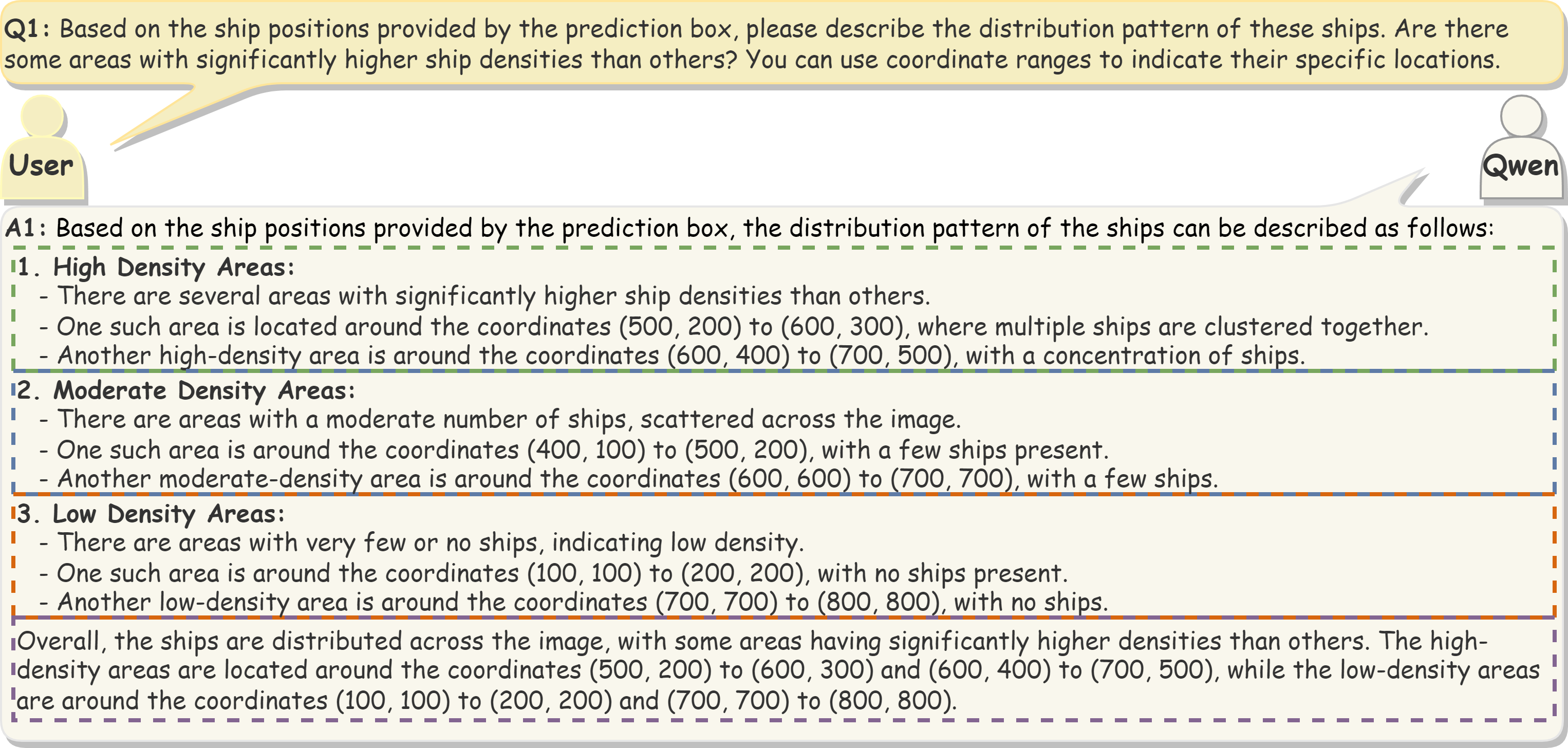}}
	\caption{Sample 2 Dialogue 1.}
	\label{fig10}
\end{figure}

In dialogue 1, we guided the VQM to analyze the information regarding the distribution of ships, as shown in Fig. \ref{fig10}, the VQM analyzes the density of ships in some areas. In order to further judge the correctness of the model's answers, we plotted the coordinates on the image, as shown in Fig. \ref{fig12} (a). It can be seen that the range of coordinates in the model’s answers matches its description, but this range could be further expanded for greater accuracy.

\begin{figure}
	\centerline{\includegraphics[scale=0.48]{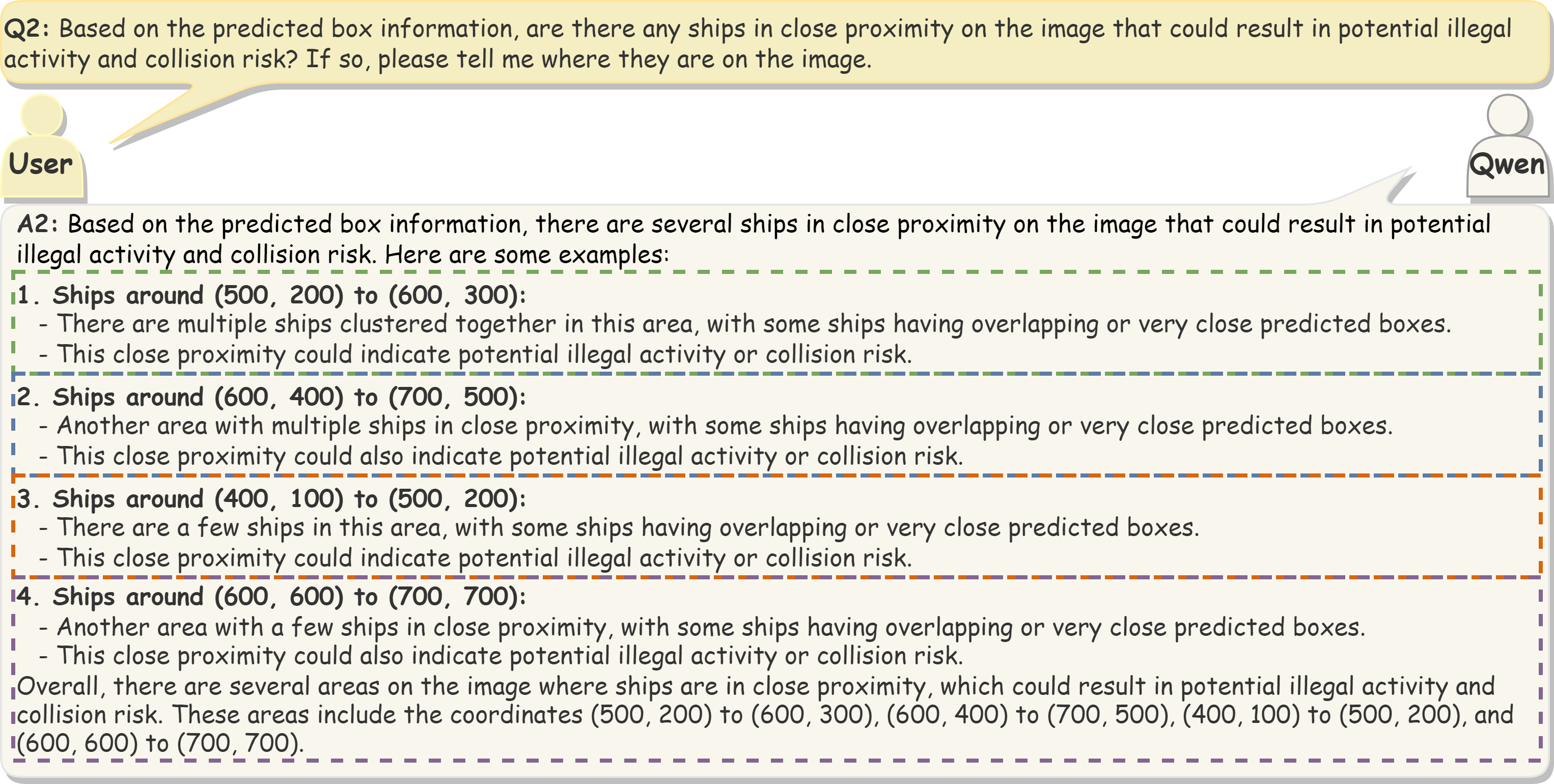}}
	\caption{Sample 2 Dialogue 2.}
	\label{fig11}
\end{figure}

As shown in Fig. \ref{fig11}, in Dialogue 2, the model is instructed to assess the presence of illegal activities and risks based on the distribution of ships, in a manner similar to Dialogue 1. We have plotted an image of their coordinates, as shown in Fig. \ref{fig12} (b).

\begin{figure}
	\centerline{\includegraphics[scale=0.24]{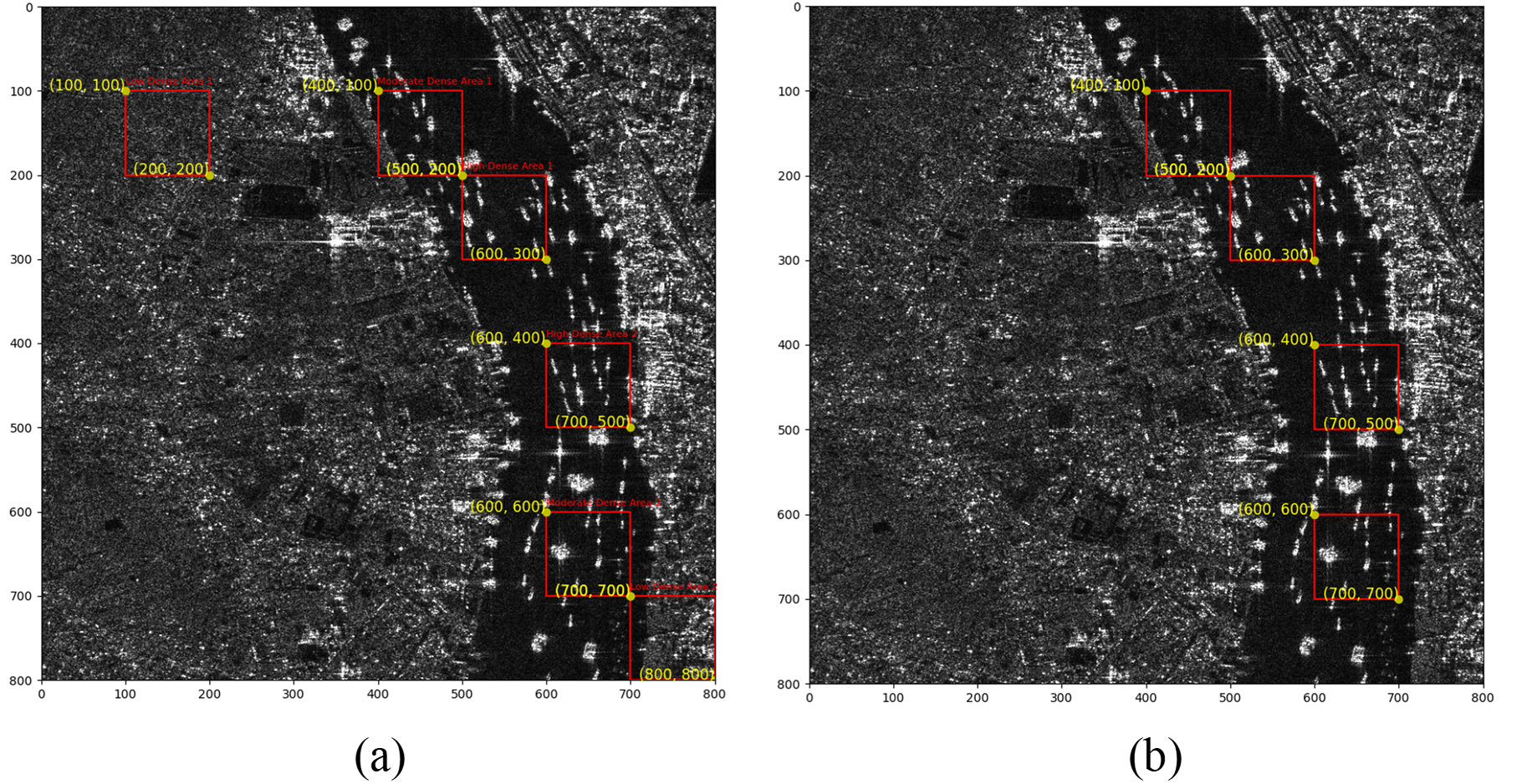}}
	\caption{Model response area visualization in Sample 2. (a) Dialogue 1. (b) Dialogue 2.}
	\label{fig12}
\end{figure}
\section{Conclusion and Future Work}

This letter introduces a method that combines target detection with visual language models for automated ship detection and Q\&A based on SAR images. By using detection outputs to guide visual language responses, this approach enables effective maritime monitoring and SAR analysis without specialized datasets or fine-tuning. Future work will focus on improving detection accuracy and refining prompts design to maximize model performance in noisy environments.

\bibliographystyle{IEEEtran}

\bibliography{ref}

\end{document}